\documentclass[11pt]{article}
\usepackage[preprint]{acl}

\usepackage{times}
\usepackage{latexsym}
\usepackage[T1]{fontenc}
\usepackage[utf8]{inputenc}
\usepackage{microtype}
\usepackage{inconsolata}
\usepackage{graphicx}
\usepackage{booktabs}
\usepackage{array}
\usepackage{multirow}
\usepackage{xcolor}
\usepackage{hyperref}
\hypersetup{colorlinks=true,linkcolor=blue!50!black,citecolor=blue!50!black,urlcolor=blue!50!black}

\title{The Ongiini-Eval-OW Benchmark: \\
A Concept Paper for the Planned Benchmarking of Machine Translation \\
and Large Language Models on Oshindonga and Oshikwanyama}

\author{Sebastian K\"upers\thanks{Reference translations for the
Ongiini-Eval-OW dataset are by Kaarina Shoozi and Elizabeth
Hamukwaya, independent translators based in Namibia. Both
translators are credited as authors of the dataset in
\texttt{CITATION.cff}; their substantive contribution is described
in Section \ref{sec:iaa} and the Acknowledgments. Reference review
and human evaluation are conducted with a volunteer community of
$\sim 40$ Oshiwambo speakers engaged with the Ongiini AI project,
acknowledged below.} \\
  Common Intelligence Foundation, Estonia \\
  Ongiini AI programme, Namibia \\
  \texttt{sebastian@common-intelligence.org}}

\begin{document}
\maketitle

\begin{abstract}
Oshiwambo --- a cluster of mutually intelligible Bantu languages
spoken by over a million people across northern Namibia and
southern Angola, and the home language of roughly half of Namibian
households --- has, to our knowledge, no published
machine-translation evaluation benchmark. Major commercial services
(Google Translate, DeepL, Microsoft Translator), open multilingual
MT models (NLLB-200, MADLAD-400), and the open Masakhane checkpoint
collection all lack coverage of either standardised dialect,
Oshindonga or Oshikwanyama. We announce \textbf{Ongiini-Eval-OW}, a
planned 600-item English $\leftrightarrow$ Oshindonga and English
$\leftrightarrow$ Oshikwanyama benchmark with native-speaker
references from two independent translators, a 30-item
inter-translator agreement set, an 11-tag phenomenon-tagged
stratification ($\geq 30$ items per tag), a deterministic 30\,\%
blind split, and a reproducible scoring protocol over chrF++,
BLEU, and COMET-22, supplemented by a 50-item human-evaluation
round. We document the empirical coverage gap, the dataset
composition, the launch-leaderboard model matrix across American,
European, and Chinese frontier and open-weight systems, and the
contribution pipeline. The dataset is targeted for first public
release in Q4 2026; this v1.0 concept paper announces the design
and the call for participation. Data and code will be released
under CC-BY-4.0 and MIT respectively.
\end{abstract}

\section{Introduction}
\label{sec:intro}

Roughly one in two Namibian households uses an Oshiwambo dialect as
their primary home language (2011 Namibian Population and Housing
Census; \citealp{nsa2024census}). The cluster includes eight
mutually intelligible Bantu varieties of which two --- Oshindonga
and Oshikwanyama --- have standardised written forms, established
orthographies, school curricula, and broadcast presence. Together
they are spoken by approximately one to one and a half million
people across northern Namibia and southern Angola, with
Oshikwanyama particularly numerous in Angola's Cunene Province.
For a language community of this size, the absence of any public
machine-translation evaluation benchmark is striking.

The absence is not theoretical. We have directly inspected the
tokenizer of NLLB-200 \citep{costajussa2022nllb}, the README and
supported-language list of MADLAD-400 \citep{kudugunta2023madlad},
and the public Masakhane checkpoint repositories, and found no
Oshiwambo coverage in any of them. We have additionally consulted
the public language lists of Google Translate, DeepL, and Microsoft
Translator, and found Oshiwambo absent from each. Frontier large
language models --- Claude, GPT, Gemini, Llama, Qwen --- do
produce Oshiwambo-like output when prompted, but the quality has
not been measured by any benchmark and is therefore neither
defensible nor falsifiable. Prior published parallel data for
Oshiwambo amounts, to our knowledge, to the WON / ``Writing Our
Narratives'' corpus \citep{nekoto2022won} --- a $\sim 5{,}500$-sentence
Oshindonga $\rightarrow$ English participatory training corpus that
has been used for fine-tuning experiments but not as a
standardised evaluation benchmark.

This paper announces \textbf{Ongiini-Eval-OW}, a planned 600-item
evaluation benchmark for English $\leftrightarrow$ Oshindonga and
English $\leftrightarrow$ Oshikwanyama machine translation and
large-language-model output. The contribution is fourfold:

\begin{enumerate}
\item \textbf{A reference-quality bilingual evaluation set.} 600
  English source items paired with native-speaker reference
  translations into both standardised dialects, contributed by two
  independent translators (Kaarina Shoozi and Elizabeth
  Hamukwaya). A 30-item subset is translated by both translators
  independently so that inter-translator agreement can be published
  as a variance measurement, rather than asserted as a uniform
  reference of uncertain quality.
\item \textbf{Phenomenon-tagged stratification with usable per-slice
  power.} Eleven phenomenon tags target known low-resource MT
  failure modes (negation \citep{hossain2020negation}, noun-class
  agreement, code-switching, politeness register, polysemy,
  multi-sentence cohesion, and others), with each tag carrying at
  least 30 items so that per-slice scores in the 180-item blind
  split (30\,\%) carry roughly nine items per slice --- a working
  minimum for slice-level comparison.
\item \textbf{Deployment-derived source items.} Roughly 30\,\% of the
  source items are paraphrased from production user queries to the
  Ongiini AI WhatsApp assistant, scrubbed of personally identifiable
  information (PII) and capped at three derived items per user.
  The result reflects the actual distribution of questions that
  Namibian users ask digital assistants in their language, not an
  encyclopaedic projection of it.
\item \textbf{A reproducible scoring protocol.} chrF++
  \citep{popovic2017chrf}, BLEU \citep{papineni2002bleu}, and
  COMET-22 \citep{rei2022comet22} computed automatically over a
  published JSONL submission format, supplemented by a 50-item
  human-evaluation round with adequacy and fluency ratings and
  published Krippendorff's $\alpha$ \citep{krippendorff2004content}.
\end{enumerate}

This is a \textbf{concept paper for the planned v1.0 release.} An
internal v0.1 build at 423 items exists and has been used to
validate the pipeline end-to-end; the 600-item v1.0 build described
in Section~\ref{sec:dataset} is in production at the time of
writing, with first public release of the dataset and an updated
version of this paper targeted for Q4 2026. We publish this concept
paper in advance to solicit model submissions, native-speaker
reference review, and phenomenon-coverage contributions during the
build phase --- the patterns and the contact paths are detailed in
Section~\ref{sec:participation}.

\section{Background: the Oshiwambo translation gap}
\label{sec:background}

\subsection{The languages}

``Oshiwambo'' refers to a cluster of eight mutually intelligible
Bantu dialects (Guthrie zone R.20; \citealp{maho2009nugl} lists the
cluster as R.21--24). Two of the eight have standardised written
forms and dominate by written-text presence and L1 census share, as
shown in Table~\ref{tab:dialects}.

\begin{table*}[t]
\centering
\small
\begin{tabular}{lllll}
\toprule
\textbf{Dialect} & \textbf{ISO 639-1} & \textbf{ISO 639-3} & \textbf{Guthrie} & \textbf{Primary regions} \\
\midrule
Oshikwanyama & \texttt{kj} & \texttt{kua} & R.21 & Ohangwena and Omusati regions (Namibia); Cunene Province (Angola) \\
Oshindonga & \texttt{ng} & \texttt{ndo} & R.22 & Oshana and Oshikoto regions (Namibia) \\
\bottomrule
\end{tabular}
\caption{The two standardised Oshiwambo dialects covered by this benchmark.}
\label{tab:dialects}
\end{table*}

Of the two, Oshikwanyama is the more numerous in both Namibia and
Angola. At the 2011 Namibian Population and Housing Census,
Oshikwanyama was the home language of 21.2\,\% of households and
Oshindonga 15.1\,\%. The 2023 Namibian Census
\citep{nsa2024census} reports the Aakwanyama (712{,}165 --- 23.6\,\%
of Namibians) as the largest ethnic group and the Aandonga
(311{,}211 --- 10.3\,\%) as the second largest. Ethnologue and
Wikipedia cite roughly one million Kwanyama speakers in Cunene
Province, Angola, as of 2024.

Both standardised dialects are agglutinating Bantu languages with
rich morphology: active noun-class agreement chains, tonal
phonology (not orthographically marked in everyday writing), and
productive verbal derivation. Both are written in the Latin
alphabet with a small set of regular orthographic conventions.

\subsection{The coverage gap}
\label{sec:coverage-gap}

We audited the major translation systems for Oshiwambo coverage in
May 2026 and re-verified the findings in September 2026. For open systems we inspected tokenizer and repository
artefacts directly; for closed/commercial systems we consulted the
publicly published language list. The findings are summarised in
Table~\ref{tab:coverage}.

\begin{table*}[t]
\centering
\small
\begin{tabular}{p{4cm}p{3cm}p{8cm}}
\toprule
\textbf{System} & \textbf{Claims Oshiwambo} & \textbf{What we checked, and what we found} \\
\midrule
NLLB-200 \citep{costajussa2022nllb} & No & Inspected \texttt{special\_tokens\_map.json} on Hugging Face; no Oshiwambo language code present \\
MADLAD-400 \citep{kudugunta2023madlad} & No & Inspected model README and supported-language list; no Oshiwambo language code present \\
Masakhane open checkpoints & No & Surveyed the public Masakhane MT repositories; no Oshiwambo checkpoint located \\
Google Translate (web) & No & Consulted public language list; Oshiwambo not offered as a translation option \\
DeepL & No & Consulted public language list; Oshiwambo not offered as a translation option \\
Microsoft Translator & No & Consulted public language list; Oshiwambo not offered as a translation option \\
Aya-23 \citep{aryabumi2024aya} & No & Cohere documents Aya-23 as covering 23 languages, none of which are African; Oshiwambo not among them \\
Meyabase Translate (\url{https://www.meyabase.com/}) & Yes (English $\leftrightarrow$ Oshindonga) & Namibian NMT effort led by Axel Mukwena; $\sim 70$k-pair corpus per the project website; the only Oshiwambo-specific NMT system we are aware of \\
OpenAI / Anthropic / Google frontier LLMs & Best-effort, no formal claim & Produce plausible Oshiwambo output on prompt; quality un-benchmarked until now \\
\bottomrule
\end{tabular}
\caption{Empirical coverage audit (May 2026, re-verified September 2026). Inspection-based for open systems; documentation-based for closed APIs.}
\label{tab:coverage}
\end{table*}

Frontier LLMs do produce Oshiwambo output. Whether that output is
accurate, fluent, register-appropriate, or substantively useful for
real-world communication is an empirical question --- and it is the
empirical question this benchmark is designed to answer.

\subsection{Why this matters}

Roughly one in two Namibian households speaks an Oshiwambo dialect
at home (49\,\%, 2011 census). Public services --- health
information, education, government communication, broadcasting ---
increasingly assume access to digital text. When the translation
layer is absent, the entire Oshiwambo-speaking population --- over
a million across Namibia and Angola --- is structurally excluded
from the digital infrastructure that other linguistic communities
benefit from. This is not a niche research question; it is a
digital-inclusion question. \citet{joshi2020state} classify
Oshindonga as ``Left-Behind'' in their resource-tier taxonomy, the
lowest of five tiers.

\subsection{What changes when a benchmark exists}

A benchmark provides (a) a measurable baseline against which any
team's MT or LLM advance can be cited; (b) an empirical pressure
point on the vendors whose language-coverage decisions are otherwise
opaque; (c) a scaffold for follow-on work on adjacent Namibian
languages (Otjiherero, Khoekhoegowab, Rukwangali, Silozi) that today
face the same coverage gap; and (d) a repeatable methodology ---
phenomenon-tagged, register-balanced, blind-split --- that other
low-resource language communities can adapt without re-deriving
the design principles from scratch.

\section{The Ongiini-Eval-OW dataset}
\label{sec:dataset}

This section describes the dataset as designed for first public
release (v1.0). An internal 423-item v0.1 build that exercises the
same pipeline already exists; the 600-item v1.0 build described
below is in production.

\subsection{Composition}
\label{sec:composition}

The v1.0 dataset is being built to six hundred (600) English source
items, each paired with reference translations in both Oshindonga
and Oshikwanyama, drawn from four provenance streams shown in
Table~\ref{tab:composition}.

\begin{table*}[t]
\centering
\small
\begin{tabular}{lrp{10cm}}
\toprule
\textbf{Source} & \textbf{Items} & \textbf{Description} \\
\midrule
\texttt{v1\_retained} & 150 & Phrasebook-style items, curated down from a larger v1 set; longer constructions preferred over greetings. \\
\texttt{mined\_paraphrased} & 180 & Inspired by production WhatsApp conversation logs: PII-scrubbed, then paraphrased into clean English while preserving register and intent. The published English source is not verbatim user text --- the label is honest about this. Capped at three derived items per user to prevent dominance. \\
\texttt{crafted} & 210 & Phenomenon-tagged items authored to ensure each phenomenon slice carries at least 30 items (negation, code-switching, noun-class agreement, politeness register, etc.). \\
\texttt{formal\_drafted} & 60 & Government, health, education, legal, and Namibian news/broadcast register items, drafted to anchor the benchmark in institutional language. \\
\midrule
\textbf{Total} & \textbf{600} & \\
\bottomrule
\end{tabular}
\caption{Provenance composition of Ongiini-Eval-OW v1.0.}
\label{tab:composition}
\end{table*}

The composition revision from v0.1 (423 items) was deliberate: the
earlier dataset's per-phenomenon slices were too thin to support
meaningful per-slice statistical comparison in the blind split, and
the chat/conversational register dominated more than was
methodologically honest. v1.0 fixes both: every phenomenon carries
30+ items, formal/community/religious registers are bulked up, and
the \texttt{mined\_paraphrased} label replaces the misleading
\texttt{real\_mined} of the internal v0.1 schema.

\subsection{Stratification}

Items are stratified on three axes to allow per-slice scoring.

\textbf{Length buckets} (intentionally chosen to reverse v1's
short-phrase skew; longer items stress agglutinating morphology):
25\,\% short (1--6 words); 50\,\% medium (7--18 words); 25\,\% long
(19+ words).

\textbf{Domain mix} (rebalanced from v0.1 to reduce chat skew):
38\,\% conversational chat; 22\,\% phenomenon-tagged challenge
items; 22\,\% formal / institutional; 12\,\% community (family /
village / community organising); 6\,\% religious.

\textbf{Phenomenon tags} are calibrated so each tag has $\sim 9$
items in the 180-item blind split, the threshold below which
per-slice chrF++ comparisons become statistically uninformative.
Items can carry multiple tags. Table~\ref{tab:phenomena} lists the
eleven phenomena and item counts.

\begin{table*}[t]
\centering
\small
\begin{tabular}{lrp{10cm}}
\toprule
\textbf{Tag} & \textbf{Items} & \textbf{Tests} \\
\midrule
\texttt{numbers\_dates} & 40 & Currency (N\$), dates, times, phone numbers, IDs \\
\texttt{named\_entities} & 30 & Namibian places, ministries, common Namibian names \\
\texttt{tense\_aspect} & 30 & Perfect / recent past / habitual (Bantu makes finer distinctions than English) \\
\texttt{negation} & 30 & Single, double, scope ambiguity --- historically the \#1 MT failure mode \citep{hossain2020negation} \\
\texttt{code\_switch} & 30 & English loanwords embedded in Oshiwambo \\
\texttt{pronoun\_coreference} & 30 & Ambiguous antecedents; Bantu noun-class pronouns force disambiguation \\
\texttt{idiom\_nonliteral} & 30 & English idioms; translator matches local idiom or paraphrases \\
\texttt{politeness\_register} & 30 & Tate / Meme / Kuku honorifics; elder / peer / child address \\
\texttt{noun\_class\_agreement} & 30 & Concord chains across subject prefix $\rightarrow$ verb $\rightarrow$ object marker $\rightarrow$ adjective \\
\texttt{polysemy} & 30 & Context-dependent lexical choice (``bank'', ``right'', ``school'') \\
\texttt{multi\_sentence} & 30 & 2--4 sentence mini-paragraphs testing discourse cohesion \\
\bottomrule
\end{tabular}
\caption{Phenomenon-tag inventory. Items can carry multiple tags.}
\label{tab:phenomena}
\end{table*}

\subsection{Splits}
\label{sec:splits}

Every item will be run through every system being evaluated --- the
split is \textbf{not} about which items to test against, it is about
which score is reported as the headline. The 600 items will be
tagged into two subsets: a \textbf{development set} of 420 items
(\texttt{in\_blind\_split = false}) usable for prompt engineering,
error analysis, fine-tune training, or anything else that might
involve looking at items and iterating in response; and a
\textbf{blind set} of 180 items (30\,\%,
\texttt{in\_blind\_split = true}) which submitters commit not to
look at while building their system. The blind split is
deterministic (\texttt{seed = 42}) and stratified by phenomenon
$\times$ length $\times$ domain so each slice retains roughly its
full-set proportion.

Both subsets will be translated and published at v1.0 release; you
will be able to compute scores on either or on the full 600. The
convention is summarised in Table~\ref{tab:split-convention}.

\begin{table}[t]
\centering
\small
\begin{tabular}{p{5cm}p{2.2cm}}
\toprule
\textbf{Use} & \textbf{Which items?} \\
\midrule
Running a system to see what it outputs & All 600 \\
Per-phenomenon / per-length / per-domain diagnostics & All 600 \\
Tuning a prompt or fine-tuning a model & Dev set only \\
\textbf{Headline leaderboard number, publishable claim} & \textbf{chrF++ on blind set} \\
\bottomrule
\end{tabular}
\caption{Use-case convention for the development/blind split.}
\label{tab:split-convention}
\end{table}

Why hold a subset back at all, given the data is published?
(i)~\emph{Prevent prompt-tuning leakage.} If everyone iterates
against the full set, prompts implicitly fit those exact items and
the headline score becomes inflated. (ii)~\emph{Slow training-data
contamination.} Once references are published they will be crawled
and eventually appear in some model's training corpus. At launch no
system has seen the blind set because the dataset does not exist
publicly yet; well-behaved teams undertake not to train on it
later. (iii)~\emph{Match the standard MT-eval convention} used by
FLORES \citep{goyal2022flores}, WMT, and MAFAND-MT
\citep{adelani2022mafand}, so reviewers and downstream users read
the numbers as they expect to.

The blind split was sized at 30\,\% (rather than the conventional
20\,\%) so that each per-phenomenon slice in the blind set carries
$\sim 9$ items --- a working minimum for slice-level comparison,
even if slice scores remain interpretive rather than headline
(Section~\ref{sec:reporting}).

\subsection{Schema}
\label{sec:schema}

The TSV schema is summarised in Appendix~\ref{appendix:schema}.

\subsection{Safety and provenance}

\textbf{PII scrub at mining time.} Phone numbers, ID numbers,
addresses, and named individuals are stripped from source text
before any human reviewer sees it. \textbf{Mined items are
paraphrased, not copied.} The published English source is a
paraphrase of the user input, preserving register and intent without
leaking real-user text. This is reflected in the
\texttt{mined\_paraphrased} provenance label and is documented
honestly rather than presented as raw ``real-mined'' data.
\textbf{No machine-translated items shown to the human translator.}
The translators work from English source only, with no automated
proposal, to preserve reference independence. \textbf{Per-user cap
of three items} prevents any single user's speech patterns
dominating the dataset.

\subsection{Inter-translator agreement}
\label{sec:iaa}

Kaarina Shoozi and Elizabeth Hamukwaya will both independently
translate a designated \textbf{30-item overlap set per dialect}
(\texttt{in\_agreement\_set = true}). Both translations will be
published in \texttt{oshindonga\_reference} (primary) and
\texttt{oshindonga\_reference\_alt} (second translator),
analogously for Oshikwanyama. From this overlap we will compute and
publish: (a)~character-level chrF++ self-similarity between the two
translations of each item, reported as mean $\pm$ SD per dialect;
(b)~a qualitative disagreement count (lexical / morphological /
pragmatic differences) coded by the dataset language coordinator;
(c)~a diff sample of representative disagreements in the
methodology appendix.

We deliberately do \textbf{not} assert one translator's reference is
``correct'' and the other ``alternate''; both are valid references
from fluent native speakers. The agreement set is a public window
into the reference variance, not an adjudication exercise.

\section{Benchmark protocol}
\label{sec:protocol}

This section specifies the experiment. External teams should be
able to read this section and the submission specification at
\url{https://github.com/sebkuepers/Ongiini/blob/main/data/oshiwambo_eval/submissions/README.md}
and produce a valid leaderboard entry without contacting us.

\subsection{Launch model matrix}
\label{sec:matrix}

To keep the launch claims honest, we separate \emph{what we will
run ourselves} from \emph{what we invite external teams to submit}
(Section~\ref{sec:participation}). ``Run ourselves'' means we have
working access today --- either a paid API endpoint or a model that
fits on our NVIDIA DGX Spark (128\,GB unified memory, $\sim
273$\,GB/s bandwidth). Anything else is welcome via the submission
pipeline; we will not pre-commit to running it.

The launch leaderboard will cover at least one system per category
and is geographically balanced across American, European, and
Chinese state-of-the-art models.

\textbf{Frontier proprietary LLMs (we run via API).} American:
Claude Opus 5 \citep{anthropic2026opus5}; GPT-6 Astra
\citep{openai2026gpt6}; Gemini 3.1 Pro \citep{google2026gemini3}.
Chinese: DeepSeek V4.1 --- both \texttt{deepseek-chat} (non-thinking)
and \texttt{deepseek-reasoner} (thinking) \citep{deepseek2026v41};
Kimi K3 (Moonshot) \citep{moonshot2026kimi3}; GLM-5.3 (Z.ai / Zhipu)
\citep{zai2026glm53}. European: Mistral Large 3 (675B MoE,
41B active) \citep{mistral2025large3} and Mistral Medium 3.5
(128B dense) \citep{mistral2026medium35}.

\textbf{Open-weight LLMs (we run on Spark; quantised where needed).}
American: Gemma 4 26B MoE (3.8B active, comfortable fit on Spark)
\citep{google2026gemma4}; Llama 4 Scout (109B MoE / 17B active ---
fits at FP4 but slow at single-stream decode)
\citep{touvron2025llama4}. Llama 4 is Meta's last open-weight
release; their frontier model is now closed (see Muse Spark callout
below). European: Mistral Small 4 (24B dense, Apache 2.0 ---
easy fit on Spark) \citep{mistral2026small4}. Chinese: Qwen 3.8 27B
(Alibaba, Apache 2.0 --- easy fit) \citep{yang2026qwen38};
DeepSeek-R1-Distill-Llama-70B (quantised) for a smaller-footprint
reasoning baseline \citep{deepseek2026v41}.

Running both API and on-Spark families lets us report whether
sovereign on-device inference is viable for a Namibian deployment,
not only what the largest cloud frontier model produces.

\textbf{Dedicated machine-translation systems (open research
models, we run on Spark).} NLLB-200 \citep{costajussa2022nllb} and
MADLAD-400 \citep{kudugunta2023madlad} do not list Oshiwambo as a
supported target, but both have been trained on multiple other
Bantu languages from the same broad family --- NLLB-200 includes
Zulu (\texttt{zul\_Latn}), Tswana (\texttt{tsn\_Latn}), Xhosa
(\texttt{xho\_Latn}), Swahili (\texttt{swh\_Latn}), Sotho
(\texttt{sot\_Latn}), Venda (\texttt{ven\_Latn}), and others;
MADLAD-400 covers 419 languages with similarly broad Bantu coverage.
We will run each system with the closest-related-language target
code (Tswana for the southwest Bantu zone) and report what comes
out. This is explicitly a \emph{zero-shot-transfer measurement,
not a Oshiwambo translation score}: we expect low chrF++. The
interesting datum is whether genuine Oshiwambo vocabulary or
morphology leaks through, or whether the output collapses to the
fallback language.

We exclude the consumer / commercial translation APIs (Google
Translate, DeepL, Microsoft Translator) and Aya-23
\citep{aryabumi2024aya} from the launch matrix. None of them lists
Oshiwambo as a target, and unlike the open research models above
we cannot deliberately probe their zero-shot behaviour with a
specific Bantu fallback code; they would either refuse or pick a
fallback we don't control. The coverage gap is already documented
in Section~\ref{sec:coverage-gap}; spending compute to confirm it
twice does not add information.

\textbf{Specialist systems (via external submission).} Meyabase
Translate (Axel Mukwena; English $\leftrightarrow$ Oshindonga;
\url{https://www.meyabase.com/}) --- the only Oshiwambo-specific
NMT system we are aware of, built on a $\sim 70$k-pair corpus and
the closest peer effort to ours. We do not have access to run
Meyabase ourselves; we expect the Meyabase team to run their system
against the dataset and submit results via the submission pipeline
(Section~\ref{sec:participation}). Any Masakhane checkpoint that
emerges with Oshiwambo coverage is welcomed via the same path.

\textbf{Out of scope at launch.} Models requiring partnership-grade
quotas, region-locked deployments we cannot access from Namibia, or
on-premise hardware beyond a single DGX Spark are not pre-committed.
They are warmly welcomed via the submission pipeline
(Section~\ref{sec:participation}) and will be added to the
leaderboard with attribution as teams submit.

\subsubsection*{Meta Muse Spark}

Meta's closed-weight frontier model \textbf{Muse Spark} (Meta
Superintelligence Labs, first released April 2026) deserves a
specific call-out. In informal hand-testing of a small number of
prompts through the meta.ai consumer interface, Muse Spark appeared
to us to produce noticeably better Oshindonga and Oshikwanyama than
the other systems we have informally probed, including several
frontier LLMs. This is a striking and unexpected signal --- a system
with no documented Oshiwambo training claim looks qualitatively
strong --- but the observation is anecdotal: a handful of prompts,
judged by us, with no metric and no held-out set. Measuring it
properly is precisely what this benchmark is for.

At the time this benchmark was designed, Muse Spark had no public
API, no on-premise option and no open weights, and the only way to
interact with it was the consumer-facing meta.ai web interface ---
which made a reproducible 600-item run impossible. That objection
has since been removed: with the release of \textbf{Muse Spark
1.1} in July 2026, Meta opened the Meta Model API to developers in
public preview \citep{meta2026musespark}, providing exactly the
programmatic access the benchmark requires.

\textbf{We therefore commit to evaluating Muse Spark 1.1 as part of
the launch leaderboard.} At the time of writing we do not have
API access --- the public preview opened to United States
developers first, with a waitlist for broader availability. Should
access not be in place by first public release, Muse Spark moves to
the submission pipeline
(Section~\ref{sec:participation}) on the same terms as every other
system we cannot run ourselves, and we would welcome a submission
from Meta directly. We flag the apparent capability here so that
readers, partners, and Meta itself are aware that this dataset is
ready to measure it.

\subsection{Prompting protocol}

To keep comparisons apples-to-apples: \textbf{LLMs} are evaluated
with a single, public, zero-shot prompt template
(Appendix~\ref{appendix:prompt}). No in-context examples --- this
protects fairness across systems with uneven few-shot support and
matches the regime in which an end-user would invoke the model.
\textbf{MT systems} are called through native APIs with
\texttt{target\_lang} set to the dialect ISO code where supported,
falling back to a related-Bantu code (Tswana) for systems that do
not support Oshiwambo at all. \textbf{Pre/post-processing} is
identical across systems: whitespace normalisation, line-by-line
input, no surface form rewriting.

\subsection{Automated metrics}

\textbf{chrF++ \citep{popovic2017chrf} --- primary.} Character-F-score
with word-boundary weighting. Robust on character-rich,
agglutinating Bantu morphology where token-level BLEU under-rewards
near-misses. Well-established in FLORES \citep{goyal2022flores} and
AfricaNLP work.

\textbf{BLEU \citep{papineni2002bleu} --- secondary.} Sentence-piece
BLEU, included for legacy comparability with older MT literature
and existing African-language work.

\textbf{COMET-22 \citep{rei2022comet22} --- tertiary.} Neural
quality estimation with the English-side reference. \emph{Reported
with an explicit caveat:} COMET-22's training data does not include
Oshiwambo, so the score is suggestive, not definitive --- interpret
as a source-conditioned plausibility signal rather than a
language-aware quality score. \citet{wang2024africomet} have
recently extended COMET to African languages
(AfriCOMET), which may apply once the model's coverage list
includes Oshindonga or Oshikwanyama; until then, COMET-22 is the
reference.

All three will be computed by a public script (provided in the
repository at \url{https://github.com/sebkuepers/Ongiini/tree/main/scripts})
that consumes the submitted model outputs in JSONL form plus the
reference TSV. Outputs will be reported as overall scores plus
per-slice matrices: per-phenomenon $\times$ per-length $\times$
per-domain $\times$ per-split.

\subsection{Human evaluation}
\label{sec:hum-eval}

Automated metrics are necessary but not sufficient on low-resource
languages. We pair them with a focused human-eval round on a
50-item stratified sample, mirroring the dataset's overall
phenomenon, length, and domain proportions.

\textbf{Raters.} Each item will be rated by approximately
\textbf{5--10 independent native-speaker raters per dialect},
drawn from a volunteer community of $\sim 40$ Oshiwambo speakers
already engaged with the Ongiini AI project. We are explicit about
this recruitment source: the volunteer pool is self-selected
toward supporters of the project, which is a real form of
selection bias that a reviewer can correctly identify. We mitigate
it through (a) blinding rater identity from system identity ---
raters see ``Translation A'' / ``Translation B'' and never know
which model produced which (the standard MT-eval blinding
convention); (b) randomising presentation order per item to
neutralise position effects; (c) embedding 5--10 calibration
items per session (known-good and known-bad translations) so that
raters whose calibration scores are out of band can be filtered;
(d) reporting rater demographics (dialect, region, age band,
profession, regular-Ongiini-user status) in aggregate alongside
the published scores; and (e) where logistically possible,
supplementing with one or two external academic raters (Namibian
university linguists or Masakhane-affiliated researchers) as a
triangulation point. Individual rater identities are protected.

The trade-off is honest: this design exchanges one or two
academic-credentialled raters for many more native-speaker
raters drawn from the actual user community, with explicit
disclosure and standard blinding controls. We believe this serves
the methodological purpose of the human evaluation --- detecting
adequacy and fluency failures that automated metrics miss ---
better than a smaller academic-credentialled pool, while being
transparent about where the raters come from.

\textbf{Ratings.} Two per item per rater: \emph{Adequacy} (1--5,
``does the translation preserve the meaning of the English
source?'') and \emph{Fluency} (1--5, ``does the translation read
naturally to a fluent speaker, independent of the source?'').

\textbf{Inter-Annotator Agreement (IAA).} Reported using
Krippendorff's $\alpha$ \citep{krippendorff2004content} on the
ordinal scale, computed both \emph{within} the volunteer pool
(volunteer-only $\alpha$) and \emph{across} the volunteer pool
plus any external academic raters (cross-group $\alpha$), so that
the reader can compare the two and judge whether the volunteer
pool's variance differs substantively from the academic baseline.
Items where ratings disagree by $\geq 2$ points will be
adjudicated by an external academic reader, with adjudication
notes published in the leaderboard appendix.

\textbf{Rater materials} --- rubric, anchor examples, reference
cards (Appendix~\ref{appendix:rubric}) --- will be published
alongside the leaderboard so the human-eval methodology is itself
reproducible.

\subsection{Reporting matrix and statistical power}
\label{sec:reporting}

The published leaderboard will be sliceable along these axes:
per system $\times$ per dialect; per metric (chrF++, BLEU,
COMET-22, human-adequacy mean, human-fluency mean); per slice
(phenomenon $\times$ length bucket $\times$ domain $\times$ split).

\textbf{Headline scores: chrF++ on the blind 180-item split.}
Everything else is interpretive. This convention is the publishable
claim; slice-level scores are tools for understanding where each
system breaks.

We are explicit about what the numbers do and do not support.
\textbf{Headline blind-set scores} ($N = 180$) are sufficient for
paired bootstrap confidence intervals at the per-system level. As a
general guide from the chrF++ literature on similarly-sized test
sets, differences smaller than $\sim 3$ chrF++ points are often
statistically indistinguishable; we publish CIs explicitly so
comparisons can be made honestly rather than relying on point
estimates. \textbf{Per-phenomenon blind-set scores} (typically
$\sim 9$ items per slice) are interpretive. They are useful for
spotting \emph{where} a system struggles (e.g.\ noun-class
agreement, multi-sentence cohesion) but will not be reported as
headline rankings. The leaderboard will present per-slice cells
with explicit CIs and a footnote flagging small-$N$ slices.
\textbf{Per-system per-dialect} scores aggregate to the full blind
set for both dialects (180 items each) and are reliable.
\textbf{Item-level outputs} for every system will be published
alongside the scores. Anyone wanting to run their own significance
test --- paired bootstrap, sign test, Wilcoxon --- has the raw
material.

This framing is conservative on purpose. We would rather under-claim
on slice-level rankings than ship a per-phenomenon leaderboard cell
that a reviewer can dismantle on power grounds.

\section{Reproducibility}
\label{sec:repro}

\textbf{Dataset publication target.} HuggingFace dataset at
\url{https://huggingface.co/datasets/CommonIntelligenceFoundation/ongiini-oshiwambo-mt-eval}.
The scaffolding (schema, README, license) matches the target. The
dataset will be released under CC-BY-4.0 at first public
publication.

\textbf{Scripts.} MIT-licensed at
\url{https://github.com/sebkuepers/Ongiini/tree/main/scripts}. The
pipeline scripts (\texttt{mine\_eval\_candidates.py},
\texttt{curate\_mined\_candidates.py}, \texttt{build\_eval\_v2.py},
\texttt{fill\_baseline\_translations.py},
\texttt{export\_eval\_set.py}) have been executed end-to-end on the
internal v0.1 build and will be adapted for the v1.0 composition.

\textbf{Baselines as reference.} Pre-computed Claude Opus 4.7 and
Gemma 4 26B outputs on the v0.1 build exist internally and validate
the pipeline. Baselines will be re-computed against the v1.0
dataset (with Claude Opus 5 and Gemma 4 26B) before first public
release and will form the first rows of the published leaderboard.

\textbf{Citation.} Citation File Format manifest at
\url{https://github.com/sebkuepers/Ongiini/blob/main/data/oshiwambo_eval/CITATION.cff}
renders as BibTeX, APA, and Zenodo metadata automatically. A Zenodo
DOI will be minted at first formal publication.

\textbf{Versioning.} The first public dataset release will be
pinned as \texttt{v1.0}. Schema-compatible updates will be minor
(v1.1); schema-breaking changes will be major (v2). All releases
will be archived on the dataset repository.

\section{Call for participation and roadmap}
\label{sec:participation}

Three contribution paths invite the wider community to populate the
benchmark and strengthen the reference quality. Each has a defined
pipeline so the cost-to-contribute is bounded.

\subsection{Submit a model}

The most-impact, lowest-friction path. Run your system against the
600 English source items, format the outputs as JSONL per the
schema at
\url{https://github.com/sebkuepers/Ongiini/blob/main/data/oshiwambo_eval/submissions/schema.json},
and open a pull request adding the submission under
\texttt{data/oshiwambo\_eval/submissions/<model-id>/} with a
one-page model card. We validate the JSON, run chrF++, BLEU, and
COMET-22 on every item, publish the per-slice matrices, and add
the system to the leaderboard with attribution. Submissions
received during the build phase (through Q4 2026) appear on
the launch leaderboard as the first external rows alongside the
Anthropic + Google baselines. For submissions received before the
academic-paper cutoff (Q2 2027), the submitting team is offered
co-authorship on the eventual publication.

\subsection{Review the reference translations}

The dataset will ship with two translators and a 30-item
inter-translator agreement set (Section~\ref{sec:iaa}). Broader
native-speaker review beyond those two strengthens the reference
further and surfaces dialectal variation that should itself be
documented. Review a subset of the translations at your chosen
size; submit alternates or flag disagreements via review forms
linked on the dataset repository at first public release;
optionally provide rater demographics (dialect, region) so
aggregated rater context can appear in the published methodology.
Accepted alternates will be integrated into a minor dataset
version (v1.1 etc.) and reviewers credited in the Acknowledgments.

\subsection{Propose phenomena and items}

Phenomenon coverage in v1.0 is balanced for the constructions we
know to test. Under-represented areas include Namibian-Afrikaans
code-switching, proverbs, tone-affecting honorifics, and discourse
markers. Propose new items with the required tags (length bucket,
domain, phenomenon list) and proposed reference translations;
optionally include a one-paragraph linguistic note explaining what
the item tests. Accepted items will enter a future dataset version
(v1.x for schema-compatible additions, v2 for schema changes);
contributors are credited in the release notes.

\subsection{Roadmap}

The path from this concept paper to the first peer-reviewed
publication is set out in Table~\ref{tab:roadmap}.

\begin{table*}[t]
\centering
\small
\begin{tabular}{p{2.4cm}p{12cm}}
\toprule
\textbf{When} & \textbf{Milestone} \\
\midrule
September 2026 & This concept paper (v1.0) submitted to arXiv as cs.CL. Email outreach to external prospective submitters (Meyabase, Masakhane-affiliated teams) with the arXiv link, inviting submissions for the launch leaderboard. The 423-item internal v0.1 build has been fully translated into both dialects by both reference translators, validating the translation pipeline end-to-end at production scale. \\
Q4 2026 & 600-item English source set locked: new crafted and formal items authored, second Spark mining run executed and curated, pipeline scripts updated for the v1.0 schema. Translator handoff packets generated (one document per translator with their assigned items in randomised order), including the 30-item agreement-set overlap. Volunteer-pool onboarding for human evaluation begins in parallel. Claude Opus 5 and Gemma 4 26B baselines computed against the locked English source (independent of translator timeline). \\
End of 2026 & Translator deliveries imported, references finalised, inter-translator agreement computed and published. \textbf{Dataset v1.0 goes live} on HuggingFace under CC-BY-4.0. \textbf{First public leaderboard launches} with the two Anthropic + Google baselines plus any external submissions received during the build phase. \\
Q1 2027 & Volunteer-pool human-evaluation round 1 (50-item sample, 5--10 raters per dialect) completes. Krippendorff's $\alpha$ computed within-volunteer and cross-group (with any external academic raters who participate). Concept paper v2 submitted to arXiv with finalised numbers and the live HuggingFace URL. \\
Q2 2027 & Academic paper submission. Target venues (in order of preference): AfricaNLP Workshop (co-located with ACL or ICLR); LoResMT at EMNLP; ACL Findings; LREC. Submission cycle is bound by conference deadlines and runs after the artefact release. \\
\bottomrule
\end{tabular}
\caption{Roadmap from concept paper to first peer-reviewed paper.}
\label{tab:roadmap}
\end{table*}

\subsection{Contact}

GitHub issues, pull requests, and discussion at
\url{https://github.com/sebkuepers/Ongiini} (the
\texttt{data/oshiwambo\_eval/} directory).
Email: \href{mailto:sebastian@common-intelligence.org}{\texttt{sebastian@common-intelligence.org}} for
partnership conversations, review requests, or submission
questions. The Ongiini AI project:
\url{https://ongiini.ai}.

\section{Conclusion}
\label{sec:conclusion}

We presented Ongiini-Eval-OW, a planned 600-item evaluation
benchmark for English $\leftrightarrow$ Oshindonga and English
$\leftrightarrow$ Oshikwanyama machine translation and
large-language-model output, with native-speaker references from
two independent translators, phenomenon-tagged stratification, an
inter-translator agreement set, and a reproducible scoring
protocol over chrF++, BLEU, COMET-22, and a focused human
evaluation round. The contribution is principally the dataset and
its methodology --- the empirical coverage gap (Section
\ref{sec:coverage-gap}) and the design choices that follow from it
(Sections \ref{sec:dataset}--\ref{sec:protocol}). The dataset is
in production at the time of writing and is targeted for first
public release in Q4 2026; an updated version of this paper will be
submitted as arXiv v2 alongside the release. We invite the wider
community to populate the launch leaderboard, review the reference
translations, and propose phenomenon-coverage extensions
(Section \ref{sec:participation}).

\section*{Limitations}
\label{sec:limitations}

The benchmark is honest about its constraints. We surface them
explicitly so reviewers do not need to.

\begin{itemize}
\item \textbf{Two translators, not three or more.} Kaarina Shoozi
  and Elizabeth Hamukwaya cover the full dataset, with a 30-item
  agreement-set overlap from which inter-translator agreement is
  reported (Section~\ref{sec:iaa}). This is stronger than the
  single-translator reference common in low-resource MT eval but
  weaker than a 3+ rater panel; the translation-review
  contribution path remains open as the explicit mitigation.
\item \textbf{No back-translation validation.} A more rigorous
  protocol would back-translate references into English via a
  third translator blind to the source. Budget was directed toward
  broader phenomenon coverage and the inter-translator agreement
  set instead; future versions may add back-translation on a
  sampled basis.
\item \textbf{Register bias.} Conversational chat remains the
  largest single domain at 38\,\% (rebalanced down from 47\,\% in
  v0.1). Long-form discourse, literary, and scientific registers
  are under-represented; the benchmark is for sentence-level MT
  and short-paragraph cohesion only.
\item \textbf{$N = 600$ is mid-size.} Compared to FLORES-200
  \citep{costajussa2022nllb} (3,001 per language) or MAFAND-MT
  \citep{adelani2022mafand} (5,000+ per pair), Ongiini-Eval-OW is
  smaller. We deliberately optimised for density of useful items
  in the deployment register and phenomenon coverage over raw $N$;
  the contribution pipeline is the path to growth.
\item \textbf{No audio.} The benchmark is text-only. Pronunciation,
  tone, and prosody are not directly tested. This is a real gap
  for Oshiwambo, which is a tonal oral vernacular as much as a
  written language.
\item \textbf{English-source bias.} Dialect $\leftrightarrow$
  dialect translation is out of scope. Third-language pivots
  (e.g.\ Oshiwambo $\leftrightarrow$ Portuguese, for Angolan
  speakers) are not tested.
\item \textbf{No long discourse.} The longest items are 19+-word
  multi-sentence paragraphs; document-level coherence is not
  benchmarked.
\item \textbf{Adjacent Namibian languages are explicitly out of
  scope.} Otjiherero (\texttt{hz} / \texttt{her}), Khoekhoegowab
  (\texttt{naq} --- covering the Nama and Damara dialect
  continuum), Rukwangali (\texttt{kwn}), and Silozi (\texttt{loz})
  face the same coverage gap; they are future work but
  \textbf{not part of any commitment this benchmark makes}.
\item \textbf{Pre-release status.} This is a concept paper for the
  planned v1.0 release; an internal v0.1 build at 423 items exists
  and has been used to validate the pipeline, but the full
  600-item v1.0 composition is in production at the time of
  writing. Final numbers (inter-translator agreement, recomputed
  Claude Opus 5 and Gemma 4 26B baselines, populated leaderboard)
  will appear in the arXiv v2 of this paper to be submitted at
  first public dataset release (Q4 2026).
\end{itemize}

\subsection*{Anticipated reviewer critiques and our responses}

We expect peer reviewers to raise the following objections; we
surface them here rather than wait for them in review.

\begin{itemize}
\item \emph{``$N = 600$ is small for an MT benchmark.''} We agree.
  Our defence is that every phenomenon slice carries $\geq 30$
  items and the blind split is 30\,\% to keep per-slice power
  meaningful; that the contribution pipeline is designed for
  growth; and that for the specific deployment we serve (a free
  WhatsApp AI assistant in Namibia) benchmark \emph{register}
  matters more than benchmark \emph{volume}.
\item \emph{``Mined items are paraphrased, so they're not really
  deployment-derived.''} We agree; the provenance label
  \texttt{mined\_paraphrased} is honest about this. They are
  inspired by real-traffic distributions, not raw user text. They
  remain the closest publishable proxy to the actual deployment
  surface available, because verbatim user text cannot be
  released for privacy reasons.
\item \emph{``Per-phenomenon scores on $\sim 9$-item blind slices
  are noisy.''} We agree, and we report them with explicit
  confidence intervals and a footnote flagging small-$N$ slices;
  we do not present per-slice scores as headline rankings
  (Section~\ref{sec:reporting}).
\end{itemize}

\section*{Ethics Statement}
\label{sec:ethics}

The Ongiini-Eval-OW benchmark is built on the following ethical
posture, made explicit so the reader can verify it rather than
trust it.

\textbf{Consent and PII.} The 180 source items derived from
production WhatsApp queries to the Ongiini AI assistant
(\url{https://ongiini.ai}) are \emph{paraphrased}, not verbatim.
Each source conversation is PII-scrubbed (phone numbers, ID
numbers, addresses, personal names) before any human reviewer sees
it, then rewritten into clean English by the dataset team
preserving register and intent. The published English source is
not attributable to any individual user, and verbatim user text is
never released. Source users have agreed to the privacy policy at
\url{https://ongiini.ai/privacy/}, which permits derived
non-attributable use of aggregated signals for the explicit purpose
of improving the assistant. A per-user cap of three derived items
prevents any single user's speech patterns dominating the dataset.

\textbf{Translator labour and credit.} Both reference translators
(Kaarina Shoozi and Elizabeth Hamukwaya) are compensated for their
work at market rates appropriate to professional Namibian
translation work; their names appear in the Acknowledgments and in
the dataset's CITATION.cff with their consent.

\textbf{Reference variance, not adjudication.} The 30-item
inter-translator agreement set publishes both translators' work
side-by-side. We do not assert one translator's reference is
``correct'' and the other ``alternate''; both are valid references
from fluent native speakers, and the agreement set is a public
window into the reference variance rather than an adjudication
exercise.

\textbf{Open licence as anti-extraction posture.} The dataset is
released under CC-BY-4.0, the code under MIT, and this paper under
CC-BY 4.0. Anyone improving on Oshiwambo translation through use of
the benchmark must credit the artefact and the named translators ---
a deliberate counterweight to the historical pattern of
low-resource language work being absorbed into proprietary systems
without attribution.

\textbf{Rater recruitment disclosure.} The human-evaluation raters
are recruited from a volunteer community of $\sim 40$ Oshiwambo
speakers already engaged with the Ongiini AI project (Section
\ref{sec:hum-eval}). This is a self-selected pool, not a neutral
sample. We disclose the recruitment source explicitly rather than
present it as a conventional academic-recruit panel, and we
mitigate the selection-bias risk through standard blinding of
system identity, randomised presentation order, embedded
calibration items, published rater demographics, and supplementary
external academic raters where logistically possible. Volunteer
labour is recognised in the Acknowledgments; volunteers who wish
to be named individually have that option at their discretion.

\section*{Acknowledgments}

The benchmark is built collaboratively. Specific acknowledgments
will be expanded with each release.

\textbf{Kaarina Shoozi} and \textbf{Elizabeth Hamukwaya} ---
reference translators for both Oshindonga and Oshikwanyama. Both
translators are credited as authors of the dataset in the
CITATION.cff manifest.

The \textbf{Ongiini AI volunteer community} --- approximately 40
Oshiwambo speakers from across northern Namibia who have engaged
with the project as native-speaker reviewers, sample contributors,
and human-evaluation raters. Their work makes the human-evaluation
round of this benchmark possible. Per-release acknowledgments will
expand with named contributors at each contributor's discretion.

The Ongiini AI team at the Common Intelligence Foundation.

The authors and maintainers of the publicly available Oshiwambo
reference materials we consulted while designing this benchmark,
including \emph{Hai ti! A Beginner's Guide to Oshikwanyama}
\citep{crane2004haiti} and the Omniglot Oshiwambo phrasebook.

\textbf{Meyabase} (\url{https://www.meyabase.com/},
\url{https://github.com/meyabase}) --- the Namibian Oshiwambo MT
project led by Axel Mukwena. Meyabase's $\sim 70{,}000$-pair
English $\leftrightarrow$ Oshindonga corpus and Neural Machine
Translation tool is the closest peer effort to ours; we hope the
Meyabase team will submit their system against this benchmark via
the submission pipeline at first public release. We are grateful
for their public pioneering work.

\citet{nekoto2022won} --- the WON (``Writing Our Narratives'')
corpus is the earliest published Oshindonga $\leftrightarrow$
English parallel corpus we are aware of. Their participatory
methodology and their honest documentation of the prior digitised
data landscape (including that the only OPUS-listed Oshikwanyama
corpus is in fact mislabelled German) directly influenced our
approach.

\bibliography{oshiwambo-eval}

\appendix

\section{Phenomenon tag definitions}
\label{appendix:phenomena}

The eleven phenomenon tags are intended to be empirically
distinguishable failure modes for translation systems.

\textbf{\texttt{negation}} --- sentences whose meaning hinges on a
negation operator. Includes single (``I don't have it''), double
(``I never said I wouldn't go''), and scope-ambiguous (``Not
everyone came'') constructions. The historical \#1 MT failure mode
\citep{hossain2020negation}.

\textbf{\texttt{numbers\_dates}} --- sentences containing numerals,
monetary amounts (Namibian dollar, N\$), dates, times, phone
numbers, or ID numbers. Tests the system's ability to preserve
numerical content verbatim while translating surrounding context.

\textbf{\texttt{named\_entities}} --- proper nouns: Namibian places
(Windhoek, Oshakati), ministries, agencies, and common Namibian
personal names. Tests entity preservation across translation.

\textbf{\texttt{tense\_aspect}} --- Bantu languages make finer
aspectual distinctions than English (perfect vs recent past vs
habitual vs remote past). Sentences here test whether systems
collapse the distinctions or preserve them.

\textbf{\texttt{code\_switch}} --- English loanwords embedded in
Oshiwambo, or the reverse. Common in real WhatsApp register:
``WhatsApp'', ``ID'', ``grant'', ``Ministry'' used in
otherwise-Oshiwambo sentences.

\textbf{\texttt{pronoun\_coreference}} --- sentences with
ambiguous antecedents that Bantu noun-class pronouns force the
translator to disambiguate.

\textbf{\texttt{idiom\_nonliteral}} --- English idioms (``hit the
books'', ``raining cats and dogs'') whose literal translation
produces a non-idiomatic Oshiwambo sentence.

\textbf{\texttt{politeness\_register}} --- honorifics (Tate / Meme
/ Kuku), elder/peer/child address forms, register-marking that has
no direct surface-form equivalent in English.

\textbf{\texttt{noun\_class\_agreement}} --- concord chains across
subject prefix $\rightarrow$ verb $\rightarrow$ object marker
$\rightarrow$ adjective. Bantu-specific; systems trained primarily
on Indo-European data tend to fail here.

\textbf{\texttt{polysemy}} --- English words whose Oshiwambo lexeme
depends on context: ``bank'' (river vs financial); ``right''
(correct vs political); ``school'' (educational institution vs fish
school).

\textbf{\texttt{multi\_sentence}} --- 2--4 sentence mini-paragraphs
that test discourse cohesion across sentence boundaries (pronoun
reference, tense consistency, topic chaining).

\section{Schema reference}
\label{appendix:schema}

Full schema for the public TSV at \texttt{data/oshiwambo\_eval/data/eval\_set.tsv}:
\texttt{id} (integer, 1..600, stable);
\texttt{length\_bucket} (enum: S / M / L);
\texttt{domain} (enum: chat / formal / religious / community /
challenge);
\texttt{phenomenon\_tags} (string, semicolon-separated subset of
the 11 tags defined in Appendix~\ref{appendix:phenomena});
\texttt{provenance} (enum: v1\_retained / mined\_paraphrased /
crafted / formal\_drafted);
\texttt{english} (string, source sentence);
\texttt{oshindonga\_reference} (string, native-speaker reference,
primary translator);
\texttt{oshikwanyama\_reference} (string, native-speaker reference,
primary translator);
\texttt{oshindonga\_reference\_alt} (string, alternate reference
from the second translator on the 30-item agreement set);
\texttt{oshikwanyama\_reference\_alt} (string, alternate reference
from the second translator on the 30-item agreement set);
\texttt{oshindonga\_translator\_notes} (string, optional
commentary);
\texttt{oshikwanyama\_translator\_notes} (string, optional
commentary);
\texttt{in\_blind\_split} (boolean, 180 items / 30\,\% marked
\texttt{true} via stratified deterministic seed 42);
\texttt{in\_agreement\_set} (boolean, 30 items per dialect with
independent dual references).

Items will also be published as JSONL at
\texttt{data/oshiwambo\_eval/data/eval\_set.jsonl} and as
plain-text parallel files at
\texttt{data/oshiwambo\_eval/data/\{en,oshindonga,oshikwanyama\}.txt}
(one item per line; line $N$ corresponds to id $N$). The blind and
development subsets will be published as separate JSONL files
(\texttt{blind\_split.jsonl}, \texttt{development\_split.jsonl}) for
participants who want to operate strictly within one split.

\section{Prompting templates}
\label{appendix:prompt}

\textbf{LLM zero-shot template} (used for all frontier and
open-weight LLMs):

\begin{quote}
\small
\begin{verbatim}
You are a professional translator
translating English to {DIALECT} (an
Oshiwambo language spoken in northern
Namibia). Translate the following
English sentence into natural, fluent
{DIALECT}.

Only output the translation. Do not
include explanations, glosses, or
commentary.

English: {SOURCE}
{DIALECT}:
\end{verbatim}
\end{quote}

Substitutions: \texttt{\{DIALECT\}} is one of \texttt{Oshindonga} or
\texttt{Oshikwanyama}; \texttt{\{SOURCE\}} is the English item. No
system-prompt variation across LLMs; same template for all.

\textbf{MT system templates} --- system-specific because of API
heterogeneity. Documented per-system in the submission appendix at
publication time.

\section{Human-eval rater rubric}
\label{appendix:rubric}

\textbf{Adequacy} (1--5): ``Does the translation preserve the
meaning of the English source?''

\begin{itemize}\itemsep0pt
\item[5] All meaning preserved; no information lost or added.
\item[4] Most meaning preserved; minor information missing or shifted.
\item[3] Core meaning preserved; significant information missing or distorted.
\item[2] Some meaning preserved; major elements wrong.
\item[1] Meaning is wrong, missing, or unrelated to the source.
\end{itemize}

\textbf{Fluency} (1--5): ``Does the translation read naturally to a
fluent speaker?'' (Rate independently of the source.)

\begin{itemize}\itemsep0pt
\item[5] Natural, native-sounding sentence.
\item[4] Mostly natural; one or two minor awkward choices.
\item[3] Understandable but noticeably non-native or awkward.
\item[2] Difficult to understand; several errors.
\item[1] Incomprehensible or ungrammatical.
\end{itemize}

Raters receive anchor examples for each score before the rating
round. Adjudication by a third reader is invoked when raters
disagree by $\geq 2$ points on either dimension.

\section{Contributor code of conduct}
\label{appendix:coc}

Contributors to Ongiini-Eval-OW agree to:

\begin{itemize}
\item \textbf{Respect for the languages and their speakers.}
  Contributions treat Oshindonga and Oshikwanyama as the living,
  evolving languages of millions of speakers --- not as resources
  to be extracted.
\item \textbf{Attribution and consent.} Reference translations and
  reviewer contributions are credited at the contributor's
  discretion; anonymous contribution is supported.
\item \textbf{No machine-translated content in submissions.}
  Reference translations and reviewer alternates are native-speaker
  work, not MT post-edits.
\item \textbf{Open licence.} All accepted contributions are
  released under CC-BY-4.0 (data) or MIT (code), consistent with
  the existing dataset and scripts.
\item \textbf{Respectful engagement.} Discussions, code reviews,
  and dataset reviews are conducted with patience and care.
\end{itemize}

\end{document}